\documentclass[pdflatex,sn-mathphys-num]{sn-jnl}

\usepackage{graphicx}%
\usepackage{multirow}%
\usepackage{amsmath,amssymb,amsfonts}%
\usepackage{amsthm}%
\usepackage{mathrsfs}%
\usepackage[title]{appendix}%
\usepackage{xcolor}%
\usepackage{textcomp}%
\usepackage{manyfoot}%
\usepackage{booktabs}%
\usepackage{algorithm}%
\usepackage{algorithmicx}%
\usepackage{algpseudocode}%
\usepackage{listings}%
\usepackage{caption}   
\usepackage{amsmath}    
\usepackage{multirow}
\usepackage{subcaption}
\usepackage{float}

\theoremstyle{thmstyleone}%
\theoremstyle{thmstyletwo}%

\theoremstyle{thmstylethree}%

\begin{document}

\title[Article Title]{\textbf{CATCH}: \textbf{C}omplementary \textbf{A}daptive \textbf{T}oken-level \textbf{C}ontrastive Decoding to Mitigate \textbf{H}allucinations in LVLMs}


\author[1,2]{\fnm{Zhehan} \sur{Kan}}\email{kzh24@mails.tsinghua.edu.cn}

\author[3]{\fnm{Ce} \sur{Zhang}}\email{cezhang@cs.cmu.edu}

\author[4]{\fnm{Zihan} \sur{Liao}}\email{zihanliao@berkeley.edu}

\author[4]{\fnm{Yapeng} \sur{Tian}}\email{yapeng.tian@utdallas.edu}

\author*[1,2]{\fnm{Wenming} \sur{Yang}}\email{yang.wenming@sz.tsinghua.edu.cn}

\author[1]{\fnm{Junyuan} \sur{Xiao}}\email{xiao-jy24@mails.tsinghua.edu.cn}

\author[5]{\fnm{Xu} \sur{Li}}\email{2021902007@chd.edu.cn}

\author[2]{\fnm{Dongmei} \sur{Jiang}}\email{jiangdm@pcl.ac.cn}

\author[2]{\fnm{Yaowei} \sur{Wang}}\email{wangyw@pcl.ac.cn}

\author[1,2]{\fnm{Qingmin} \sur{Liao}}\email{liaoqm@tsinghua.edu.cn}

\affil*[1]{\orgname{Tsinghua University}}

\affil[2]{\orgname{Pengcheng Laboratory}}

\affil[3]{\orgname{Carnegie Mellon University}}

\affil[4]{\orgname{University of California, Berkeley}}

\affil[4]{\orgname{The University of Texas at Dallas}}

\affil[5]{\orgname{Chang'an University}}



\abstract{Large Vision-Language Model (LVLM) systems have demonstrated impressive vision-language reasoning capabilities but suffer from pervasive and severe hallucination issues, posing significant risks in critical domains such as healthcare and autonomous systems. Despite previous efforts to mitigate hallucinations, a persistent issue remains: visual defect from vision-language misalignment, creating a bottleneck in visual processing capacity. To address this challenge, we develop \textbf{C}omplementary \textbf{A}daptive \textbf{T}oken-level \textbf{C}ontrastive Decoding to Mitigate \textbf{H}allucinations in LVLMs (\textbf{CATCH}), based on the Information Bottleneck theory. CATCH introduces Complementary Visual Decoupling (CVD) for visual information separation, Non-Visual Screening (NVS) for hallucination detection, and Adaptive Token-level Contrastive Decoding (ATCD) for hallucination  mitigation. CATCH addresses issues related to visual defects that cause diminished fine-grained feature perception and cumulative hallucinations in open-ended scenarios. It is applicable to various visual question-answering tasks without requiring any specific data or prior knowledge, and generalizes robustly to new tasks without additional training, opening new possibilities for advancing LVLM in various challenging applications.}




\maketitle

\section{Introduction}\label{sec1}

Large Vision-Language Models (LVLMs) have achieved significant advancements in areas such as visual question answering~\cite{antol2015vqa} and embodied intelligence~\cite{gupta2021embodied}, owing to their remarkable potential to integrate and interpret both visual and linguistic information. However, hallucinations in LVLMs, referring to the generation of textual content that is inconsistent with the visual input, remain a pervasive issue. Therefore, substantial risks are posed, particularly in high-stakes domains such as healthcare~\cite{chen2024huatuogpt} and autonomous systems~\cite{cui2024survey}, where erroneous decisions potentially lead to severe consequence.

Hallucinations in LVLMs primarily arises from an excessive dependence on training data, limited real-world comprehension, and a over-reliance on linguistic information due to the Large Language Model (LLM)-centric architecture of the existing models in vision-language reasoning~\cite{bai2024hallucination}. Despite the prevalence of hallucination issues across all existing LVLMs, research dedicated to mitigating this problem remains scarce. One of the primary contributors to hallucination is the quality of data. Recent efforts have focused on addressing this issue by introducing negative data~\cite{alayrac2022flamingo}, counterfactual data~\cite{yu2024hallucidoctor}, and reducing noise and errors within existing datasets~\cite{wang2024mitigating, yue2024less}. Given the imbalance in vision-language reasoning introduced by the LLM-centric architecture, some approaches have sought to enhance the model's visual reasoning capabilities by increasing resolution~\cite{chen2024internvl, liu2024improved, liu2024visual, zhai2023halle} or incorporating more advanced vision encoders~\cite{he2024incorporating, jain2024vcoder, tong2024eyes}.
Furthermore, several methods have optimized decoding strategies. For instance, HALC~\cite{chen2024halc} employs Grounding DINO to resample images, thereby enhancing LVLMs' perceptual sensitivity to fine-grained targets. Other approaches, such as VCD~\cite{leng2024mitigating} and M3ID~\cite{favero2024multi}, introduce contrastive decoding, which involves comparing the original image with either a noise-added version or text-only input, aiming to reduce the models' vulnerability to language priors. 

\begin{figure*}[t]
\centering
\includegraphics[width=1\columnwidth]{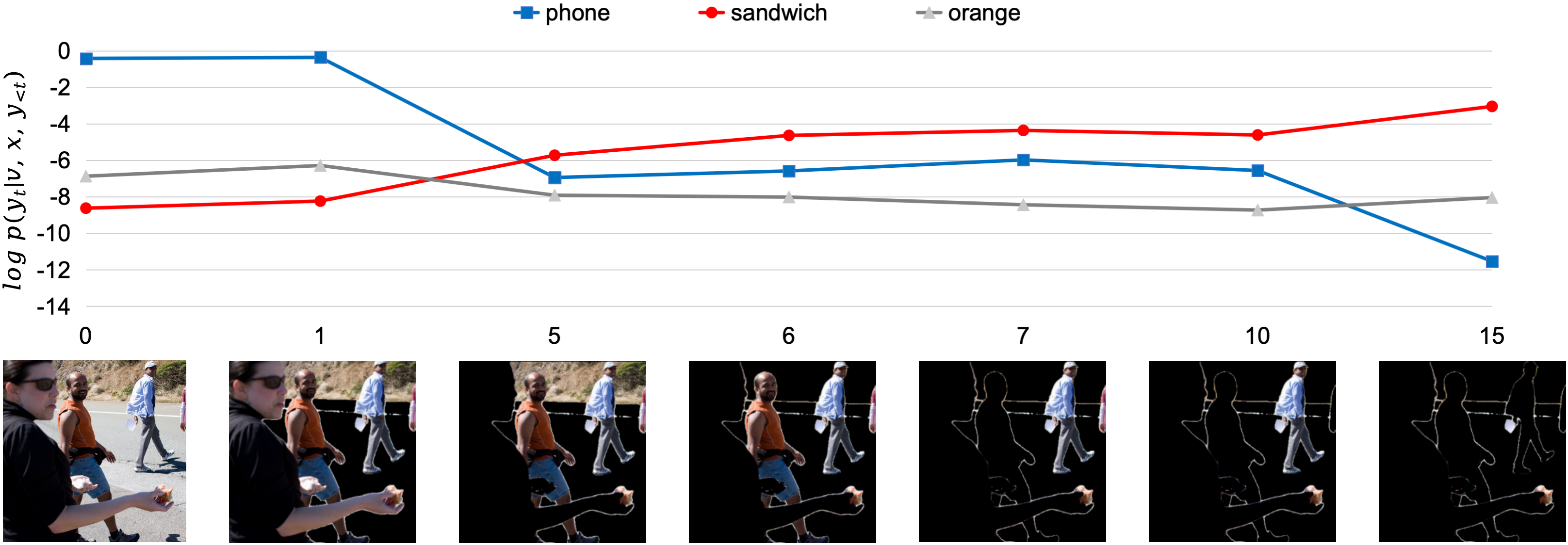}
\centering
\caption{\textbf{Analysis of how varying decoupling levels affect ground-truth token probability.} We utilize SAM to segment the original visual input into seven levels. The horizontal axis represents the number of segmented objects selected utilizing SAM, and the vertical axis represents the token probability. As irrelevant visual features unrelated to the target reduce, the probability of the hallucinated token \textit{``phone"} decreases, while the probability of the ground-truth token \textit{``sandwich"} increases.}
\label{fig:intro1}
\end{figure*}

Previous research typically attributes the causes of hallucinations to two primary factors: (1) $\textbf{statistical bias}$, unbalanced object distribution and biased object correlations in textual information, (2) $\textbf{language bias}$, overlook visual evidence and overly exploit language priors for decision-making~\cite{leng2024mitigating}. However, we observe that hallucinations in LVLMs arise from an inability to comprehensively process visual information, rendering them unaccounted for by the two factors mentioned above. We first illustrate this with the example as shown in Fig. \ref{fig:intro1}, we decouple visual input to seven levels by utilizing the Segment Anything Model (SAM)~\cite{kirillov2023segment} to segment the original visual input with different numbers of target objects. The LVLM has generated the ongoing response, \textit{``One person on the left side is holding a ..."}. Initially, it hallucinates by predicting \textit{``phone"} instead of \textit{``sandwich"}. As irrelevant visual features are reduced, the probability of generating the ground-truth token \textit{``sandwich"} increases, while the probability of the hallucinated token \textit{``phone"} decreases. Once irrelevant visual features are minimized, the probability of the ground-truth \textit{``sandwich"} significantly surpasses that of the hallucinated \textit{``phone"}. 

This phenomenon indicates that LVLMs fail to perform precise vision-language reasoning on an entire image when extraneous information (e.g., segmented portions) is present. We define the aforementioned phenomenon as \textbf{visual defect}. The visual defect primarily arises from an overload of visual information exceeding the model's visual reasoning capacity, causing disruption and uncertainty that bias reasoning toward linguistic information. In addition, visual defects worsen in open-ended scenarios as bias toward linguistic information propagates and accumulates with each token generation step, akin to noise in chaotic systems, making it increasingly difficult to retain critical visual information and thereby impeding precise reasoning. We posit that the visual defect arises from the alignment between visual and linguistic feature spaces. Specifically, visual features possess higher dimensionality than textual features, and mapping this higher-dimensional space to a lower-dimensional aligned vision-language latent space introduces an \textbf{information bottleneck}, compressing features and impeding full information propagation. This issue is intrinsic to vision-language tasks, where alignment between visual and linguistic information is necessary.
Therefore, we believe that simply optimizing high-quality data or strengthening the visual encoder is insufficient to resolve the visual defect.

\begin{figure*}[t]
\centering
\includegraphics[width=1\columnwidth]{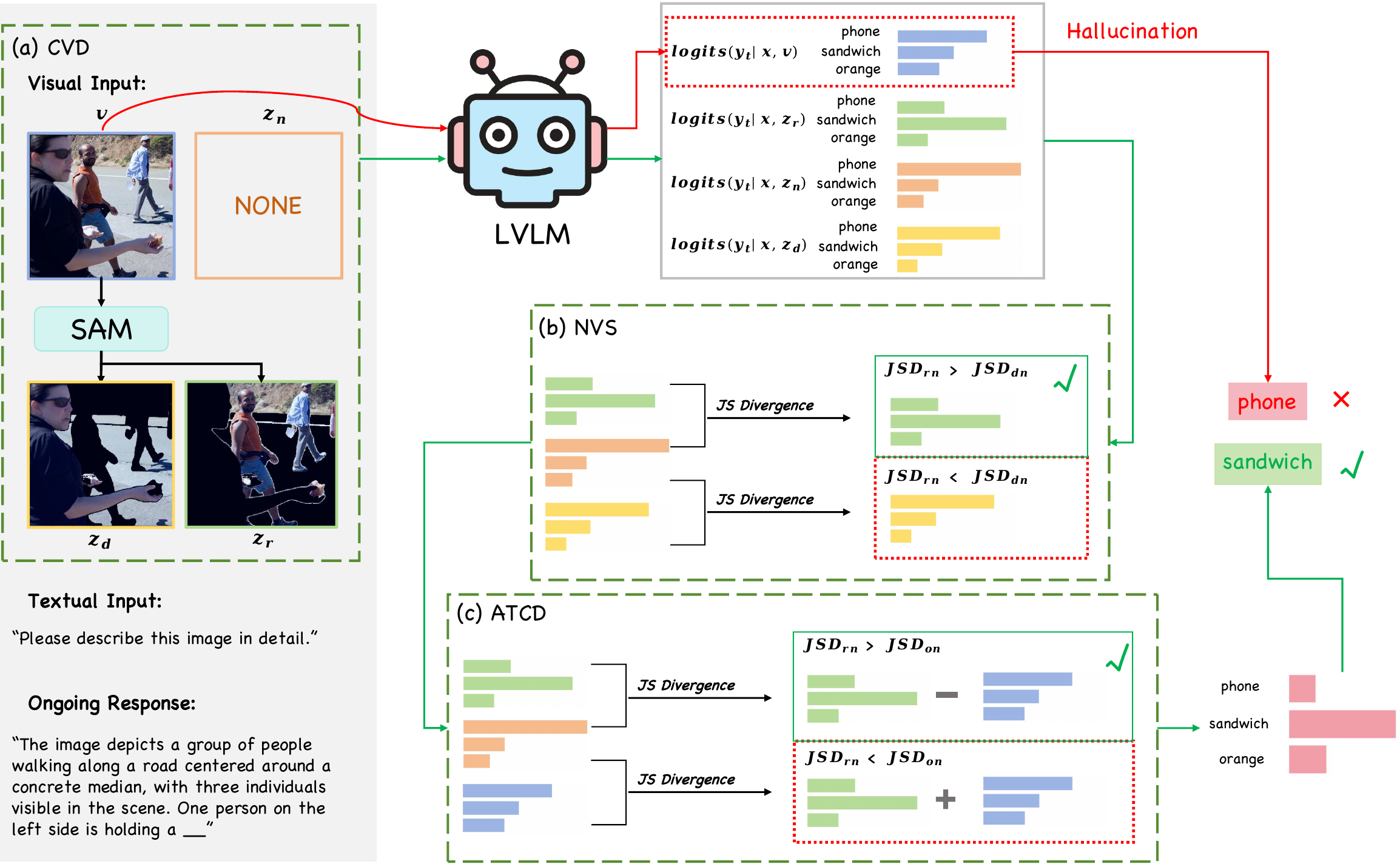}
\centering
\caption{LVLMs may generate responses that include hallucinations (e.g., \textit{``One person on the left side is holding a \underline{phone}''}, where \textit{``sandwich''} is hallucinated as \textit{``phone''}. First, the CVD method leverages SAM to decouple the original input image $v$ into the dual image $z_d$ and the residual image $z_r$, and introduces a non-visual input $z_n$. These four inputs are then passed into the LVLM to generate their corresponding output distributions: $logits_o$, $logits_d$, $logits_r$ and $logits_n$. The Jensen-Shannon Divergence (JSD) is computed between them to obtain $JSD_{on}$, $JSD_{mn}$, and $JSD_{cn}$. The NVS method compares $JSD_{mn}$ and $JSD_{cn}$, and the input with the greater distance is selected as the decoupled image (e.g., $z_r$). Next, ATCD selects the decoding strategy by comparing $JSD_{cn}$ and $JSD_{on}$, if $JSD_{cn}$ is greater, the decoupled image output distribution is employed to contrastively subtract the original distribution. Conversely, if $JSD_{on}$ is greater, the output distribution from the decoupled image is leveraged to contrastively enhance the weighted original distribution. Effectively correcting the hallucinated token (e.g., \textit{``phone''} is successfully corrected to \textit{``sandwich''}). Notably, this process is dynamically performed at each token generation step.}
\label{fig:pipeline}
\end{figure*}

We develop a method for detecting and mitigating hallucinations arising from the visual defect by introducing \textbf{C}omplementary \textbf{A}daptive \textbf{T}oken-level \textbf{C}ontrastive Decoding to Mitigate \textbf{H}allucinations in LVLMs (\textbf{CATCH}). 
To separate extraneous information and create a stable decoupled visual representation at each generation step, we propose Complementary Visual Decoupling (CVD). As illustrated in Fig. \ref{fig:pipeline}(a), before the visual input $v$ is fed into the LVLM, we utilize SAM to segment it into two complementary parts: the dual image $z_d$ and the residual image $z_r$. 
Leveraging their complementary nature, CVD divides the entire set of visual features into two simplified parts. At each generation step, the least important visual features for the next token are dynamically highlighted in one part and obscured in the other, designating the decoupled image $z$ as the highlighted one that preserves key visual features while removing extraneous details.

\begin{figure}[H]
    \centering
    \begin{subfigure}{\textwidth}
        \centering
        \includegraphics[width=1\textwidth]{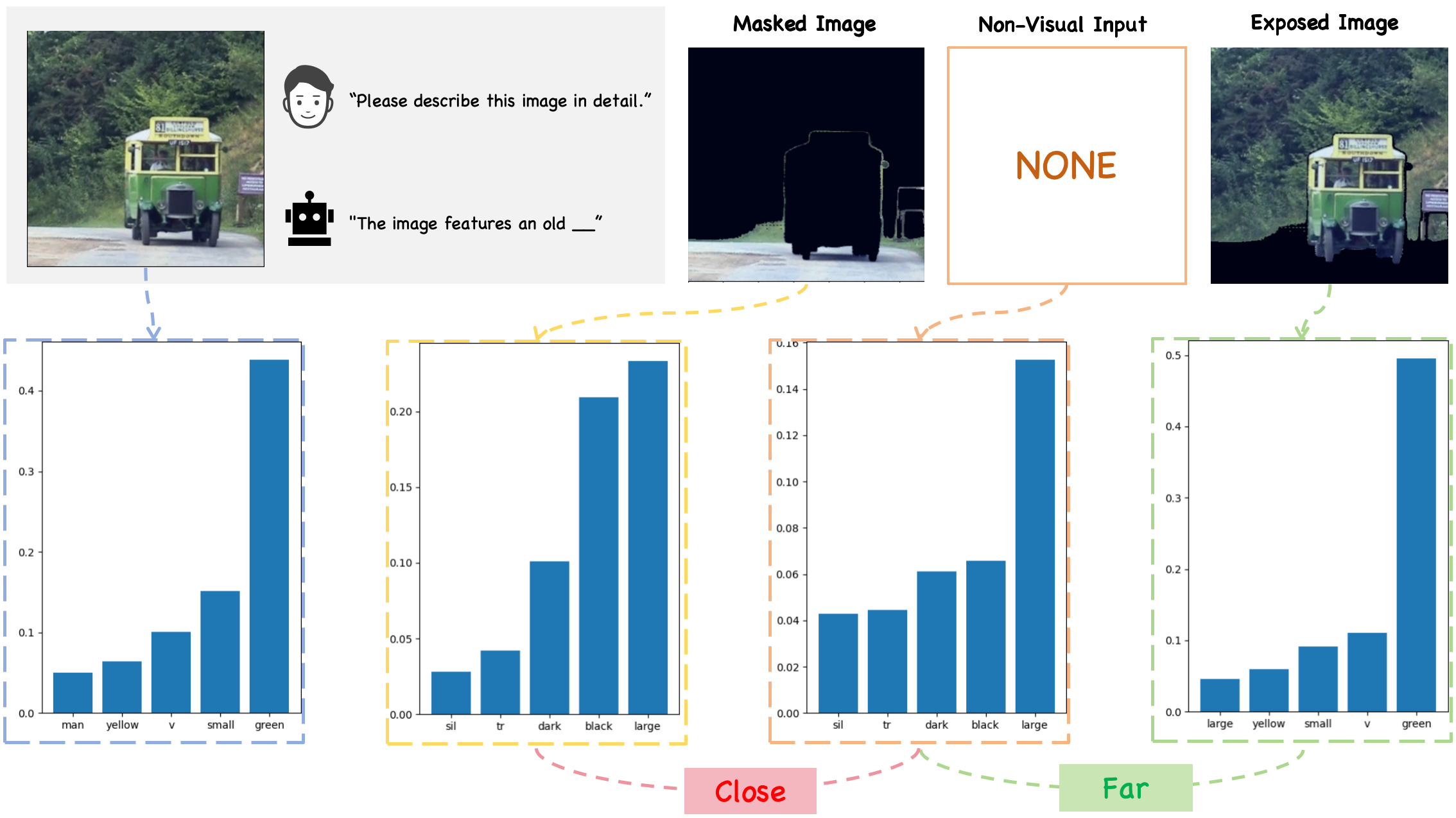}
        \caption{In the single-sample analysis, the output distribution of the masked image is nearly identical to that of the non-visual input, while the output probability of the ground-truth token from the exposed image is significantly higher compared to that of the original visual input (e.g., the ground-truth token \textit{``green"}).}
        \label{one-sample}
    \end{subfigure}
    
    \begin{subfigure}{\textwidth}
        \centering  
        \includegraphics[width=1\textwidth]{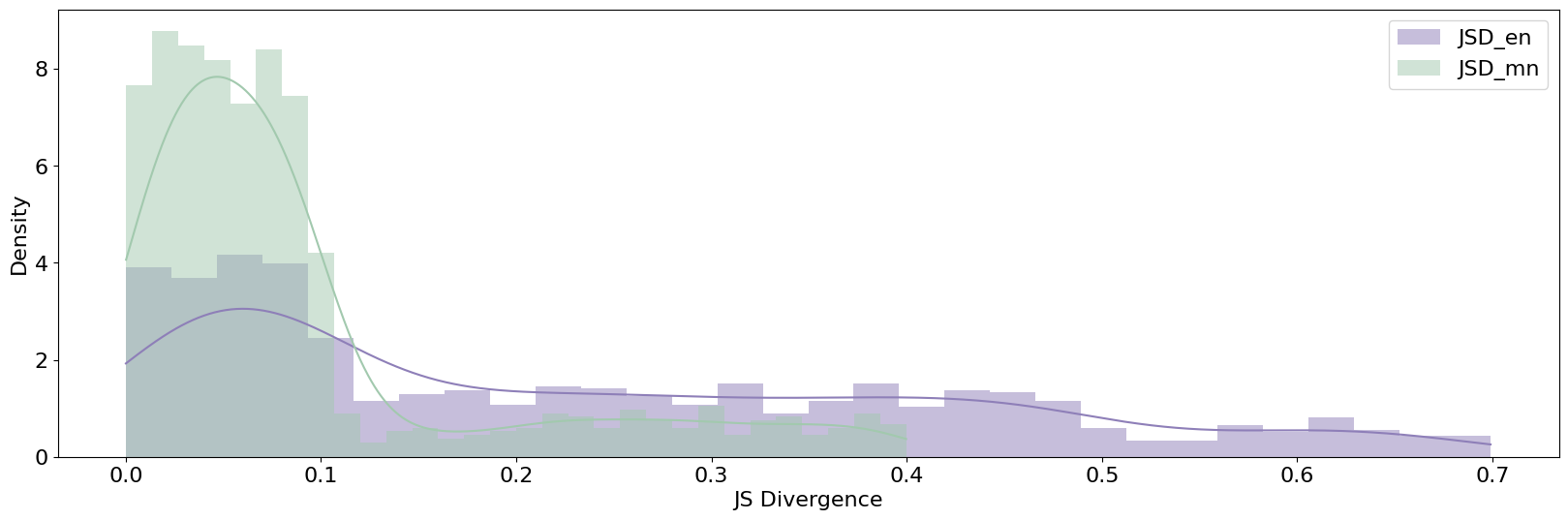}
        \caption{In the extensive statistical analysis, $JSD_{en}$ is significantly greater than $JSD_{mn}$.}
        \label{statistical analysis}
    \end{subfigure}
    
    \caption{We randomly selected 1,000 instances from the MSCOCO~\cite{lin2014microsoft} dataset, masking key visual features to create a masked image and a exposed image and then calculated $JSD_{en}$ between the exposed image and the non-visual input, $JSD_{mn}$ between the masked image and the non-visual input. We conducted a single-sample experiment as shown in (\ref{one-sample}) and performed an extensive statistical analysis as presented in (\ref{statistical analysis}).}
    \label{fig:intro2}
\end{figure}

The key question is how to identify the correct decoupled image $z$ within the set \{$z_d$, $z_r$\}. As illustrated in Fig. \ref{fig:intro2}, 1,000 instances are randomly selected from the MSCOCO~\cite{lin2014microsoft} dataset, with key visual features masked to create a masked image and a complementary exposed image.
We introduce a non-visual input $z_n$, containing no visual information, generates responses based solely on the textual prompt and generated text tokens. The Jensen-Shannon Divergence (JSD) of the output distributions between the non-visual input and both the masked image and the exposed image is then calculated, denoted as $JSD_{mn}$ and $JSD_{en}$, respectively.
From the single-sample (\ref{one-sample}) and extensive statistical analysis (\ref{statistical analysis}), we observe that: (1) the output distribution of the masked image is nearly identical to that of the non-visual input, with $JSD_{en}$ significantly greater than $JSD_{mn}$, (2) the output probability of the ground-truth token from the exposed image is much higher than that of the visual input (e.g., the token \textit{``green"}). 
The above observation indicates that when key visual features relevant to the current token are obscured, the generation process relies almost entirely on linguistic priors. Furthermore, the exposed image demonstrates that the visual decoupling method (CVD) effectively increases visual information density and reduces uncertainty associated with linguistic knowledge. 

Based on the aforementioned analyses, we propose Non-Visual Screening (NVS), as shown in Fig. \ref{fig:pipeline}(b). we introduce the non-visual input $z_n$ alongside the dual image $z_d$ and residual image $z_r$, which are fed into the LVLM to generate corresponding output distributions. We then calculate the distance between the output distributions from the non-visual input and the dual image, denoted as the \textit{dual-to-non distance}, and the distance between the output distributions from the non-visual input and the residual image, denoted as the \textit{residual-to-non distance}. The visual input corresponding to the greater value between the \textit{dual-to-non distance} and the \textit{residual-to-non distance} is identified as the decoupled image.

We consider two scenarios: (1) \textbf{Hallucination Existence}: when the distance between the output distributions from the non-visual input and the decoupled image,  denoted as the \textit{decoupled-to-non distance} is greater than the distance between the output distributions from the non-visual input and the original visual input,  denoted as the \textit{original-to-non distance}, it suggests that the output distribution from the original visual input is closer to that of the non-visual input. This indicates that the original visual input contains redundant information, leading to visual uncertainty and the amplification of language priors, ultimately causing hallucinations. (2) \textbf{Diversity Insufficient}: when the \textit{decoupled-to-non distance} is smaller than \textit{original-to-non distance}, it suggests that the output distribution from the decoupled image is closer to that of the non-visual input, indicating an insufficient global receptive field or the presence of abstract concepts in the next token. Consequently, the decoupled image lacks diversity, leading to cumulative hallucinations, which occur when reasoning becomes dominated by language priors, causing the probability of hallucinations to increase as the sequence lengthens in open-ended generation scenarios.

Based on these two scenarios, we propose Adaptive Token-level Contrastive Decoding (ATCD) as shown in Fig. \ref{fig:pipeline}(c) to contrastively mitigate hallucinations and enhance diversity at each generation step. When the first scenario occurs, the output distribution from the decoupled image is employed to contrastively subtract the original distribution, which contains hallucinated concepts. When the second scenario occurs, the output distribution from the decoupled image is leveraged to contrastively enhance the weighted original distribution, thereby improving the diversity of generation and preventing cumulative hallucinations.

CATCH can be seamlessly integrated into various LVLMs and applied across various visual question-answering scenarios without additional training or expert intervention. Our method is evaluated with LLaVA-1.5~\cite{liu2024visual} and InstructBLIP~\cite{dai2023instructblipgeneralpurposevisionlanguagemodels} as baselines, both utilizing Vicuna 7B~\cite{chiang2023vicuna} as their language decoder. The evaluation is performed on the public hallucination assessment datasets POPE~\cite{li2023evaluating}, CHAIR~\cite{rohrbach2018object}, and MME~\cite{fu2023mme}. On the POPE dataset, our method improved Accuracy and F1 score by up to 8.07 and 5.98 points, respectively, compared to the baselines. On the MME dataset, across the four subtasks—Existence, Count, Position, and Color—our method outperformed the baselines by 16\%, demonstrating CATCH's significant enhancement in the perception of various feature types. Additionally, on the CHAIR dataset, our method achieved a 45.8\% improvement over the baselines, indicating that CATCH effectively mitigates cumulative hallucinations in open-ended generation scenarios.

\section{Results}\label{sec2}

\begin{table*}[htbp]
\centering
\caption{\textbf{Results on POPE benchmark.} The best results are \textbf{bolded}, and the second-best are \underline{underlined}.}
\scalebox{0.8}{
\begin{tabular}{c c c c c c c c c c c}
\toprule
& \multirow{2}{*}{\textbf{Setup}} & \multirow{2}{*}{\textbf{Method}} & \multicolumn{4}{c}{\textbf{LLaVA 1.5}} & \multicolumn{4}{c}{\textbf{InstructBLIP}} \\
\cmidrule(lr){4-7} \cmidrule(lr){8-11}
&  &  & {{Acc. $\uparrow$}} & {{Prec. $\uparrow$}} & {{Rec. $\uparrow$}} & {{F1 $\uparrow$}} & {{Acc. $\uparrow$}} & {{Prec. $\uparrow$}} & {{Rec. $\uparrow$}} & {{F1 $\uparrow$}} \\
\midrule
& \multirow{5}{*}{Random} 
& \textit{base}   & 84.13 & 82.86 & 86.07 & 84.43 & 82.80 & 82.24 & 83.67 & 82.95 \\
&& VCD    & 85.37 & 83.14 & \textbf{88.73} & 85.84 & 83.93 & 84.42 & 82.67 & 83.73 \\
&& M3ID   & 86.00 & 85.11 & 87.27 & 86.18 & 84.37 & 84.62 & 84.00 & 84.31 \\
&& RITUAL & \underline{88.87} & \underline{89.23} & \underline{88.40} & \underline{88.81} & \underline{88.83} & \underline{90.48} & \underline{86.80} & \underline{88.60} \\
&& \textbf{CATCH}  & \textbf{90.43} & \textbf{93.04} & 87.40 & \textbf{90.13} & \textbf{90.17} & \textbf{92.28} & \textbf{87.67} & \textbf{89.91} \\
\cmidrule(lr){2-11}
\multirow{5}{*}{\rotatebox{90}{MS-COCO}}
& \multirow{5}{*}{Popular} 
& \textit{base}   & 80.87 & 78.23 & 85.53 & 81.72 & 75.80 & 72.74 & 82.53 & 77.33 \\
&& VCD    & 81.10 & 77.78 & \underline{87.07} & 82.16 & 77.73 & 75.43 & 82.27 & 78.70 \\
&& M3ID   & 82.83 & 79.62 & \textbf{88.27} & 83.72 & 77.30 & 74.10 & 83.93 & 78.71 \\
&& RITUAL & \underline{85.83} & \underline{84.17} & \textbf{88.27} & \underline{86.17} & \underline{81.97} & \underline{78.90} & \underline{87.27} & \underline{82.87} \\
&& \textbf{CATCH}  & \textbf{87.07} & \textbf{90.12} & 83.27 & \textbf{86.56} & \textbf{83.70} & \textbf{81.22} & \textbf{87.67} & \textbf{84.32} \\
\cmidrule(lr){2-11}
& \multirow{5}{*}{Adversarial} 
& \textit{base}   & 76.23 & 71.75 & 86.53 & 78.45 & 75.40 & 71.60 & 84.20 & 77.39 \\
&& VCD    & 75.60 & 70.78 & 87.20 & 78.14 & 76.80 & 73.62 & 83.53 & 78.26 \\
&& M3ID   & 77.70 & 73.23 & \underline{87.33} & 79.66 & 76.03 & 72.48 & 83.93 & 77.79 \\
&& RITUAL & \underline{78.80} & \underline{74.43} & \textbf{87.73} & \underline{80.54} & \underline{78.73} & \underline{74.57} & \underline{87.20} & \underline{80.39} \\
&& \textbf{CATCH}  & \textbf{83.17} & \textbf{83.10} & 83.27 & \textbf{83.18} & \textbf{79.90} & \textbf{75.82} & \textbf{87.80} & \textbf{81.37} \\
\midrule
& \multirow{5}{*}{Random} 
& \textit{base}   & 81.73 & 76.53 & 91.53 & 83.36 & 81.13 & 78.03 & 86.67 & 82.12 \\
&& VCD    & 81.83 & 75.74 & 93.67 & 83.76 & 82.00 & 79.38 & 86.47 & 82.77 \\
&& M3ID   & 83.57 & 77.86 & \underline{93.80} & 85.09 & 82.33 & 77.81 & 90.47 & 83.66 \\
&& RITUAL & \underline{85.17} & \underline{79.79} & \textbf{94.20} & \underline{86.40} & \underline{87.13} & \underline{83.92} & \underline{91.87} & \underline{87.71} \\
&&\textbf{CATCH}  & \textbf{89.63} & \textbf{88.83} & 90.67 & \textbf{89.74} & \textbf{89.43} & \textbf{86.22} & \textbf{93.87} & \textbf{89.88} \\
\cmidrule(lr){2-11}
\multirow{5}{*}{\rotatebox{90}{A-OKVQA}}
& \multirow{5}{*}{Popular} 
& \textit{base}   & 76.67 & 70.51 & 91.67 & 79.71 & 75.67 & 70.97 & 86.87 & 78.12 \\
&& VCD    & 74.70 & 68.12 & 92.87 & 78.59 & 76.50 & 71.69 & 87.60 & 78.85 \\
&& M3ID   & 76.80 & 70.20 & \underline{93.13} & 80.06 & 75.60 & 70.40 & 88.33 & 78.36 \\
&& RITUAL & \underline{78.83} & \underline{71.99} & \textbf{94.40} & \underline{81.68} & \underline{78.73} & \underline{72.83} & \underline{91.67} & \underline{81.17} \\
&& \textbf{CATCH}  & \textbf{84.63} & \textbf{80.90} & 90.67 & \textbf{85.51} & \textbf{80.90} & \textbf{74.54} & \textbf{93.87} & \textbf{83.09} \\
\cmidrule(lr){2-11}
&\multirow{5}{*}{Adversarial} 
& \textit{base}   & 67.40 & 61.78 & 91.27 & 73.68 & 68.00 & 63.08 & 86.80 & 73.06 \\
&& VCD    & 67.43 & 61.48 & 93.33 & 74.13 & \underline{70.67} & \textbf{65.24} & 88.47 & 75.10 \\
&& M3ID   & 68.10 & 61.99 & \underline{93.60} & 74.58 & 69.57 & 64.21 & 88.40 & 74.39 \\
&& RITUAL & \underline{68.57} & \underline{62.26} & \textbf{94.27} & \underline{74.99} & 70.27 & 64.15 & \underline{91.87} & \underline{75.55} \\
&& \textbf{CATCH}  & \textbf{75.47} & \textbf{69.53} & 90.67 & \textbf{78.70} & \textbf{71.90} & \underline{65.22} & \textbf{93.87} & \textbf{76.96} \\
\midrule
& \multirow{5}{*}{Random} 
& \textit{base}   & 81.23 & 75.42 & 92.67 & 83.16 & 79.93 & 76.73 & 85.93 & 81.07 \\
&& VCD    & 81.50 & 74.78 & \underline{95.07} & 83.71 & 81.83 & 79.03 & 86.67 & 82.67 \\
&& M3ID   & 82.83 & 76.64 & 94.47 & 84.62 & 80.57 & 76.77 & 87.67 & 81.85 \\
&& RITUAL & \underline{86.10} & \underline{80.30} & \textbf{95.67} & \underline{87.31} & \underline{84.87} & \underline{82.52} & \underline{88.47} & \underline{85.39} \\
&& \textbf{CATCH}  & \textbf{89.97} & \textbf{88.80} & 91.47 & \textbf{90.11} & \textbf{86.63} & \textbf{84.11} & \textbf{90.33} & \textbf{87.11} \\
\cmidrule(lr){2-11}
\multirow{5}{*}{\rotatebox{90}{GQA}}
&\multirow{5}{*}{Popular}
& \textit{base}   & 72.50 & 65.85 & 93.47 & 77.27 & 72.73 & 68.14 & 85.40 & 75.80 \\
&& VCD    & 71.57 & 64.72 & \underline{94.80} & 76.93 & 73.67 & 68.82 & 86.53 & 76.67 \\
&& M3ID   & 72.83 & 66.04 & 94.00 & 77.58 & \underline{74.57} & \underline{69.45} & 87.73 & 77.53 \\
&& RITUAL & \underline{74.80} & \underline{67.50} & \textbf{95.67} & \underline{79.15} & 74.50 & 69.17 & \underline{88.40} & \underline{77.61} \\
&& \textbf{CATCH}  & \textbf{82.97} & \textbf{78.18} & 91.47 & \textbf{84.30} & \textbf{76.93} & \textbf{71.24} & \textbf{90.33} & \textbf{79.66} \\
\cmidrule(lr){2-11}
&\multirow{5}{*}{Adversarial} 
& \textit{base}   & 67.63 & 61.68 & 93.13 & 74.21 & 69.57 & \underline{64.80} & 85.67 & 73.79 \\
&& VCD    & 67.47 & 61.38 & 94.20 & 74.33 & 69.43 & 64.76 & 85.27 & 73.61 \\
&& M3ID   & 68.13 & \underline{61.88} & \underline{94.47} & 74.78 & 68.90 & 64.06 & 86.13 & 73.47 \\
&& RITUAL & \underline{68.23} & 61.75 & \textbf{95.80} & \underline{75.10} & \underline{70.17} & 64.76 & \underline{88.47} & \underline{74.78} \\
&& \textbf{CATCH}  & \textbf{77.70} & \textbf{71.72} & 91.47 & \textbf{80.40} & \textbf{71.40} & \textbf{65.52} & \textbf{90.33} & \textbf{75.95} \\
\bottomrule
\end{tabular}
}
\label{tab:POPE}
\end{table*}

\textbf{Results on POPE.} As shown in Table \ref{tab:POPE}, we evaluate our CATCH method on the POPE dataset~\cite{li2023evaluating}. POPE assesses hallucinations as a binary classification task by asking yes/no questions about object presence (e.g., \textit{``Is there a dog in the image?''}). This benchmark aggregates data from three sources: MSCOCO~\cite{lin2014microsoft}, A-OKVQA~\cite{schwenk2022okvqa}, and GQA~\cite{hudson2019gqa}, and includes three subsets: random, popular, and adversarial, which address object prevalence and co-occurrence patterns. Each sampling setting uses 500 images per dataset, with 6 questions per image, resulting in a total of 27,000 query-answer pairs derived from the development sets. Evaluation is based on four key metrics: Accuracy, Precision, Recall, and F1 score.

We observed that our CATCH method outperforms the current best by a significant margin. Compared to the baseline, it improves Accuracy and F1 score by up to 8.07 and 5.98 points, respectively, demonstrating its effectiveness in mitigating hallucinated concepts present in the original distribution.

In addition, it is worth noting that while all methods exhibit a clear performance decline from the random to the popular setting, with a further drop in the adversarial setting, CATCH demonstrates superior performance in these more challenging scenarios. It improves accuracy by 6.65, 8.21, and 8.36 points over the baseline in the random, popular, and adversarial settings, respectively. These results reveal that more challenging tasks amplify the visual defects in LVLMs, whereas CATCH effectively reduces the density of extraneous visual information, preventing reliance on language priors. We also observed that, compared to Recall, CATCH achieved a more significant improvement in Precision, which can be attributed to its lower "yes" response ratio compared to the baseline. This suggests that CATCH is more conservative and stable when handling uncertain responses.

\begin{figure}[H]
    \centering
    \begin{subfigure}{\textwidth}
        \centering
        \includegraphics[width=1\textwidth]{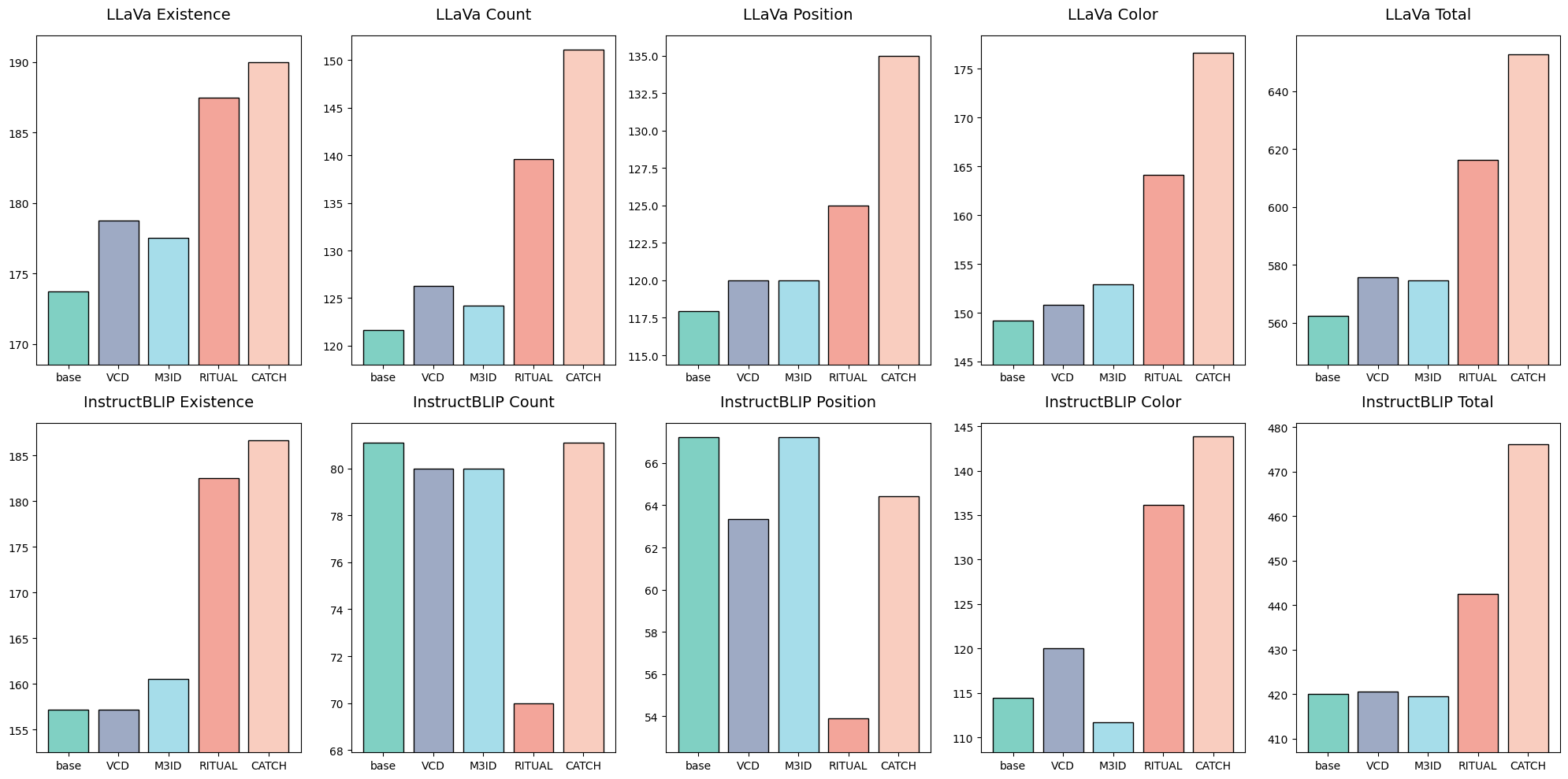}
        \caption{\textbf{Results on MME Hallucination benchmark.} For the LLaVA and InstructBLIP baselines, we evaluate the existence, count, position, and color subsets. We randomly selected five different seeds and used the average as the final result.}
        \label{fig: MME}
    \end{subfigure}
    
    \begin{subfigure}{\textwidth}
        \centering  
        \includegraphics[width=1\textwidth]{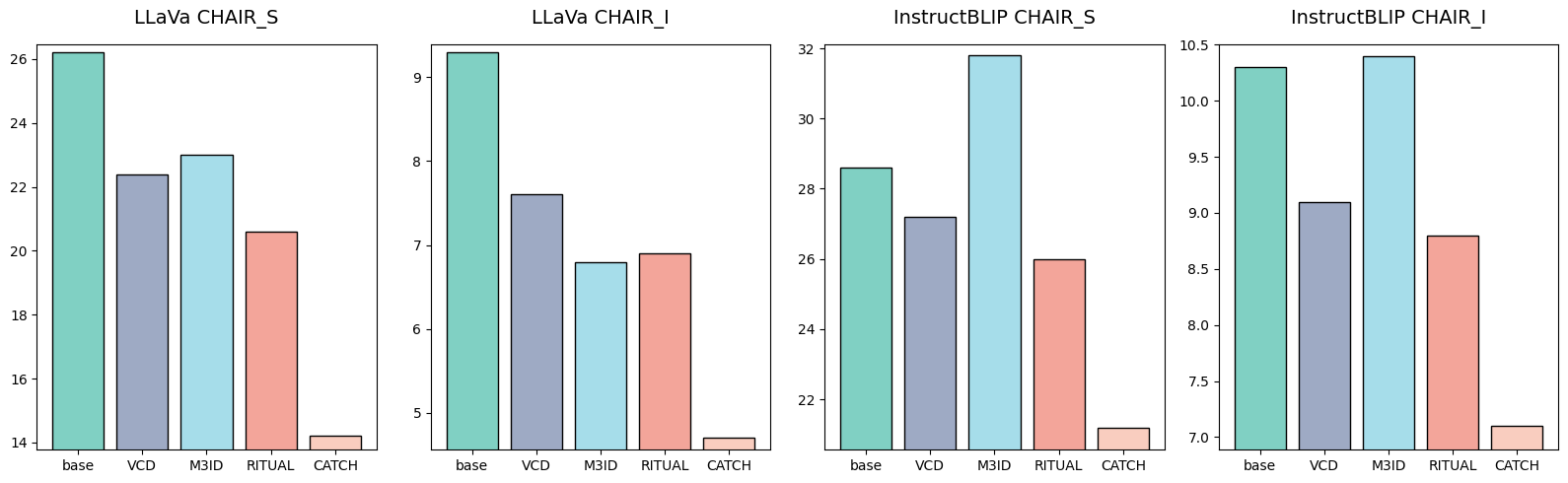}
        \caption{\textbf{Results on CHAIR benchmark.} Lower $CHAIR_S$ and $CHAIR_I$ scores indicate better performance.}
        \label{fig:CHAIR}
    \end{subfigure}  
    \caption{\textbf{Results on  MME Hallucination (\ref{fig: MME}) and CHAIR benchmark (\ref{fig:CHAIR}).}}
    \label{fig:MME and CHAIR}
\end{figure}

\textbf{Results on MME Hallucination.} MME Hallucination~\cite{fu2023mme} is a comprehensive benchmark designed for evaluating LVLMs, comprising four subsets: existence, count, position, and color. Each subset contains 30 images and 60 questions, with two questions per image. Similar to POPE, these questions are structured as binary yes/no queries, and performance is measured based on binary accuracy. 

MME Hallucination is more challenging than POPE, as it includes attribute-level hallucinations related to position and color, in addition to the existence and count dimensions.
As shown in Fig. \ref{fig: MME}, the first row presents the evaluation results with LLaVa as the baseline, while the second row shows the results with InstructBLIP as the baseline. The first four columns represent performance on the existence, count, position, and color subsets, respectively, while the final column shows the total score across all four subsets. CATCH demonstrates significant improvements in total score, with increases of 16\% and 13.4\% for LLaVa and InstructBLIP baselines, respectively. Notably, when comparing the results with LLaVa as the baseline to other LVLMs on the MME leaderboard, the positive impact of CATCH is comparable to upgrading LLaVa 7B to LLaVa 13B, GPT-4V, LLaVa 1.6 34B, and Qwen-VL-Plus across the existence, count, position, and color dimensions.

\textbf{Results on CHAIR.} CHAIR ~\cite{rohrbach2018object} uses ground-truth captions and object annotations to evaluate hallucinations in LVLMs by calculating the proportion of hallucinated objects in generated captions relative to actual objects. This evaluation is based on two metrics: $CHAIR_S = \frac{| \{ \text{hallucinated objects} \} |}{| \{ \text{all mentioned objects} \} |}, \quad CHAIR_I = \frac{| \{ \text{captions with hallucinated objects} \} |}{| \{ \text{all captions} \} |}.$ $CHAIR_S$ represents assessments at the sentence level, measuring the proportion of hallucinated sentences relative to all sentences, while $CHAIR_I$ measures hallucinations at the object instance level, indicating the proportion of hallucinated objects relative to all generated objects. Lower scores indicate fewer hallucinations. We randomly selected 500 images from the COCO validation set and conducted image captioning using the prompt, "Please describe this image in detail."
Fig. \ref{fig:CHAIR} shows the evaluation results on the CHAIR dataset. For LLaVA, CATCH achieves scores of 14.2 on $CHAIR_S$ and 4.7 on $CHAIR_I$, representing significant improvements over the baseline scores of 26.2 and 9.3, with gains of 45.8\% and 49.5\%, respectively. Similarly, when using InstructBLIP as the baseline, CATCH achieves scores of 21.2 on $CHAIR_S$ and 7.1 on $CHAIR_I$, compared to baseline scores of 28.6 and 10.3, showing improvements of 25.9\% and 31.1\%. 

\begin{figure}[H]
    \centering
    \begin{subfigure}{\textwidth}
        \includegraphics[width=1\columnwidth]{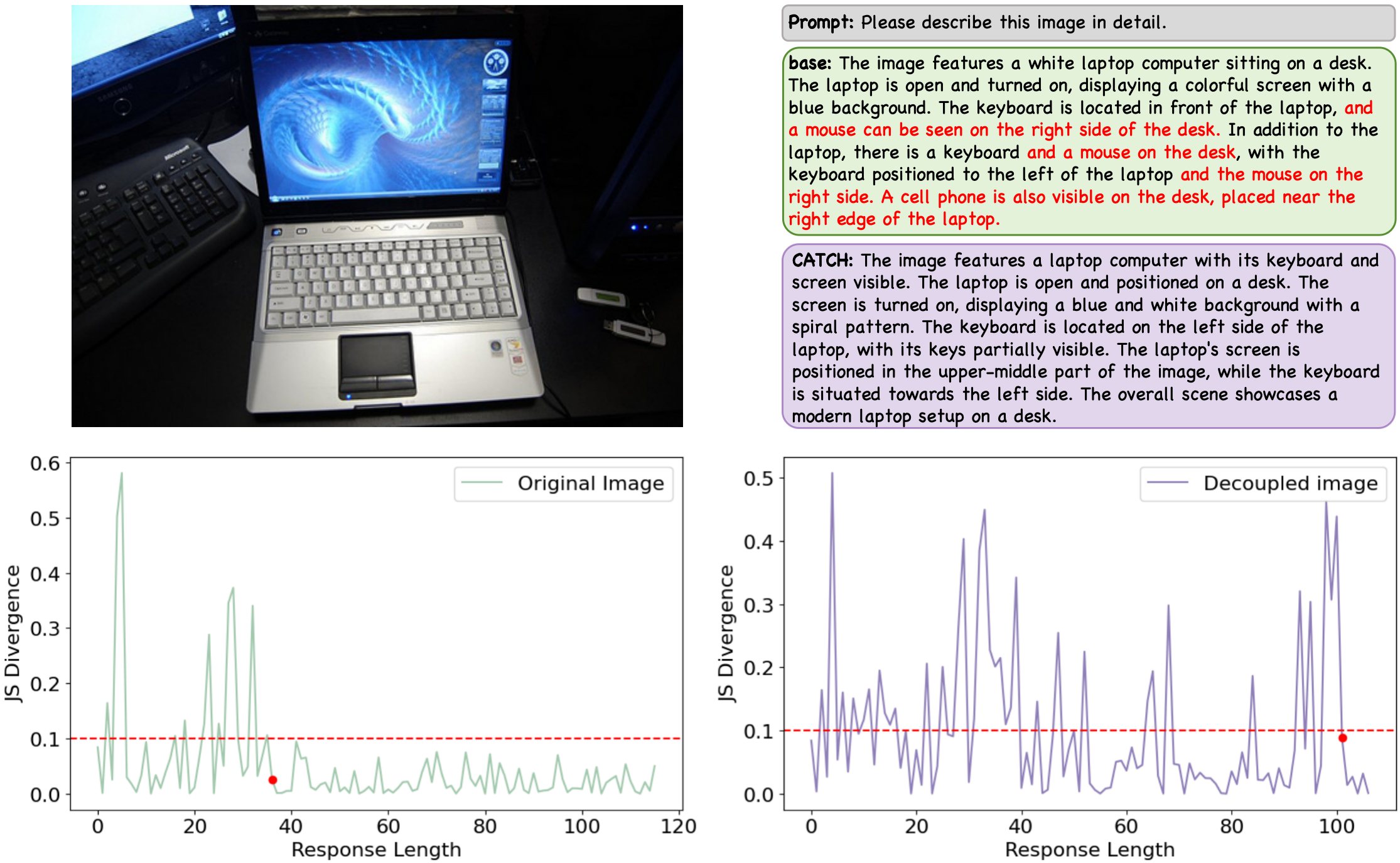}
        \centering
        \caption{The first row presents an example where CATCH eliminates hallucinations such as \textit{``mouse''} and \textit{``phone''} that were generated by the baseline. The second row shows the corresponding cumulative hallucinations analysis, indicating that the baseline model experiences cumulative hallucinations as early as the 39th token, whereas CATCH delays the occurrence of cumulative hallucinations until the 101st token.}
        \label{fig:CHAIR2}
    \end{subfigure}
    
    \begin{subfigure}{\textwidth}
        \centering  
        \includegraphics[width=1\columnwidth]{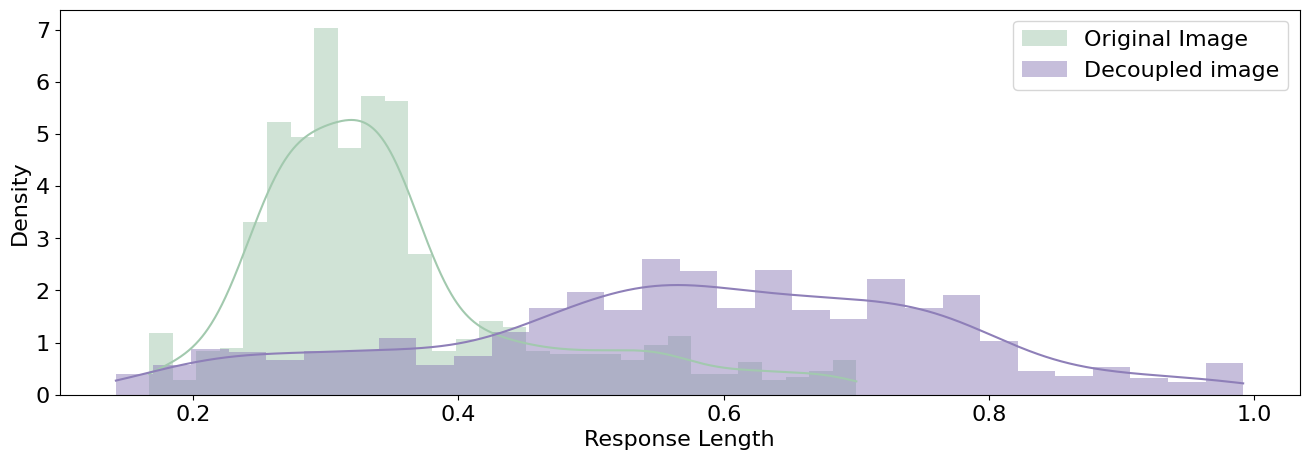}
        \centering
        \caption{We provide extensive statistical analysis across 1,000 samples, demonstrating that the baseline model experiences cumulative hallucinations for most samples by 40\% of the sentence length, whereas CATCH effectively delays cumulative hallucinations in a substantial portion of samples until approximately 80\% of the sentence length.}
        \label{fig:CHAIR3}
    \end{subfigure}
    
    \caption{An analysis of cumulative hallucinations includes a single-sample experiment, as shown in (\ref{fig:CHAIR2}), and an extensive statistical analysis, as presented in (\ref{fig:CHAIR3})}
    \label{fig:intro2}
\end{figure}

CATCH significantly prevent cumulative hallucinations. As shown in Fig. \ref{fig:CHAIR2}, we calculated the JS divergence between the output distributions from the original visual input and non-visual input with the baseline, as well as between the decoupled visual input and the non-visual input with CATCH during generation. We observed that from the 39th token onward, the output distributions from the original visual input and the non-visual input become almost identical. This indicates the onset of cumulative hallucinations, where the LVLM output relies solely on language priors, sharply increasing the likelihood of hallucinations, such as \textit{``mouse"} and \textit{``cell phone"}. In contrast, the output distribution from the decoupled image maintained a greater divergence from that of the non-visual input throughout the generation process, with cumulative hallucinations occurring only after the 101st token. This suggests that: (1) the decoupled image contains less visual uncertainty due to the separation of extraneous information compared to the original visual input, and (2) in the earlier generation steps, CATCH mitigates hallucinated concepts and enhances diversity.
For further analysis, as showm in Fig. \ref{fig:CHAIR3}, we randomly sampled 1,000 examples and plotted the distribution of the points at which cumulative hallucinations occur. The results indicate that, with the baseline, cumulative hallucinations for the original visual input mostly occur from 40\% of the sequence length, whereas for the decoupled image with CATCH, they occur mostly from 80\% of the sequence length.

\section{Methods}\label{sec11}
\subsection{Formulation of LVLMs}
We consider an LVLM parameterized by $\theta$. The model receives a textual query $x$ and a visual input $v$, where $v$ provides contextual visual information to aid the model in generating a relevant response $y$ to the query. Initially, the raw image $v$ is processed through a vision encoder to extract visual features. These features are then mapped into the input space of the large language model through a vision-language alignment module ($e.g.,$ Q-Former~\cite{li2023blip}, linear projection~\cite{liu2024visual}), generating visual tokens. Subsequently, these visual tokens, together with textual tokens obtained through embedding the query, are fed into the large language model to auto-regressively generate the response. Mathematically, this process can be formulated as follows:
\begin{equation}
    y_t \sim p_\theta \left( y_t \mid v, x, y_{<t} \right) \propto \exp  logit_\theta \left( y_t \mid v, x, y_{<t} \right),
\end{equation}
where $y_t$ denotes the token generated at time step $t$, and $y_{<t}$ represents the sequence of tokens generated up to time step $t-1$. During the decoding phase in LVLMs, hallucinations arise when the generated probability distribution deviates from the factual information provided by the visual input $v$. 
To mitigate hallucinations arising from visual defect, we propose the Complementary Visual Decouplin (CVD) for visual information separation, Non-Visual Screening (NVS) for hallucination detection, and Adaptive Token-level Contrastive Decoding (ATCD) for hallucination mitigation.  

\subsection{Complementary Visual Decoupling}
The visual defect fundamentally arises from the unbalanced alignment in vision-language multimodal integration, leading to a visual information bottleneck. Inspired by the information bottleneck principle, which addresses this issue by removing irrelevant information and preserving only information pertinent to the current prediction to create a more stable representation, we propose the Complementary Visual Decoupling (CVD) method to preserve essential visual information while removing extraneous details.

Assuming we have raw information $v$ and label $y$, the core idea of the information bottleneck principle is to map the observation of $v$ to a robust representation $z$, which retains the essential characteristics needed for predicting $y$ while simultaneously minimizing redundant information. Theoretically, treating $v$, $z$, and $y$ as three random variables, the optimization objective is to maximize the mutual information between $z$ and $y$, while minimizing the mutual information between $z$ and $v$. This optimization objective can be defined as:
\begin{equation}
    min[I(v;z) - I(z;y)].
\end{equation}
In this paper, the raw information $v$ represents the visual input, and the label $y$ represents the accurate response. We employ the Segment Anything Model (SAM) to optimize this objective. Specifically, as illustrated in Fig. \ref{fig:pipeline}, a pre-trained SAM is utilized to segment all objects:
\begin{equation}
    \{O_1, O_2, ..., O_N, B\} = SAM(v),
\end{equation}
where $O$ denotes the objects and $B$ represents the background.
Using the area of the target region as the confidence score, we select the top $M$ objects as the exposed portion and mask the remaining objects and background to obtain the dual image $z_d$. Similarly, we mask the top $M$ objects and treat the remaining objects and background as the target, generating a residual image $z_r$:
\begin{equation}
    z_d = (\sum_{i}^{M} O_i)\odot v \quad z_r = (1 - \sum_{i}^{M} O_i)\odot v.
\end{equation}
Through CVD, the original visual information is decoupled into two simplified components: the dual image $z_d$ and the residual image $z_r$, leveraging their complementary nature.

\subsection{Non-Visual Screening}
At each generation step, key visual features for the next token are dynamically emphasized in one part while being obscured in the other. The decoupled image $z$ is designated as the one that retains the highlighted key visual features, effectively eliminating extraneous details. We observed that when critical visual information relevant to the next token is obscured, the output distribution becomes dominated by language priors. Thus, to identify the correct decoupled image $z$ between the dual image and the residual image, we introduce a non-visual input $z_n$, containing only the textual prompt, without any visual information, to serve as an assistant. We then calculate the Jensen-Shannon Divergence (JSD) as the distance between the output distributions from the non-visual input and the dual image as $d(z_d, z_n)$, and between the non-visual input and the residual image as $d(z_r, z_n)$:
\begin{align}
    d(z_d,\, z_n) &= D_{JS}\left(\,p_\theta(\,y_t\,|\, z_d,\, x,\, y_{<t}) \parallel p_\theta(\,y_t\, |\, z_n,\, x,\, y_{<t})\right), \nonumber \\
    d(z_r,\, z_n) &= D_{JS}\left(\,p_\theta(\,y_t\, |\, z_r,\, x,\, y_{<t}) \parallel p_\theta(\,y_t\, |\, z_n,\, x,\, y_{<t})\right),  
\end{align}
where $D_{JS}(P || Q) = \frac{1}{2} D_{KL}(P || M) + \frac{1}{2} D_{KL}(Q || M)$, $M = \frac{1}{2} (P + Q)$, and $D_{KL}(P || Q) = \sum_{i} P(i) \log \frac{P(i)}{Q(i)}$. The visual input corresponding to the greater distance is selected as the decoupled image $z$, formulated as:
\begin{align}
    z &= 
    \begin{cases}
        z_d, & \text{if }\, d(z_d,\, z_n)\, \geq\, d(\,z_r,\, z_n); \\
        z_r, & \text{if }\, d(z_d,\, z_n)\, <\, d(\,z_r,\, z_n).
    \end{cases}
\end{align}

\subsection{Adaptive Token-level Contrastive Decoding}
We first calculate the distance between the output distributions from the non-visual input and the decoupled image, denoted as $d(z, z_n)$, and the distance between the output distributions from the original visual input and the non-visual input, denoted as $d(v, z_n)$:
\begin{align}
    d(z,\, z_n) &= D_{JS}\left(\,p_\theta(\,y_t\,\, |\, z,\, x,\, y_{<t}) \parallel p_\theta(\,y_t\, |\, z_n,\, x,\, y_{<t})\right), \nonumber \\
    d(v,\, z_n) &= D_{JS}\left(\,p_\theta(\,y_t\,\, |\, v,\, x,\, y_{<t}) \parallel p_\theta(\,y_t\, |\, z_n,\, x,\, y_{<t})\right), 
\end{align}
We consider two scenarios: (1) \textbf{Hallucination Existence}: When $d(z, z_n)$ is greater than $d(v, z_n)$, we conclude that hallucinations are present in the original output distribution. Therefore, we use the output distribution from the decoupled image to contrastively subtract the original distribution. (2) \textbf{Diversity Insufficient}: When 
$d(z, z_n)$ is less than $d(v, z_n)$, we consider there to be a risk of cumulative hallucinations. In this case, we use the output distribution from the decoupled image to contrastively enhance the weighted original distribution, thereby improving the diversity of generation, formulated as:
\begin{align}
    y_t \sim p_\theta(\,y_t\, |\, z,\, v,\, x) = 
    \begin{cases}
        \text{softmax}\left[\alpha \cdot \text{logit}_\theta(\,y\, |\, z,\, x) -\text{logit}_\theta(\,y\, |\, v,\, x)\right], \\
        \quad \quad \quad \quad \quad \quad \quad \quad \quad \quad\text{if } d(z_d,\, z_n) \geq d(v,\, z_n); \\
        \text{softmax}\left[\beta \cdot \text{logit}_\theta(\,y\, |\, v,\, x) + \text{logit}_\theta(\,y\, |\, z_d,\, x)\right], \\
        \quad \quad \quad \quad \quad \quad \quad \quad \quad \quad\text{if } d(z_d,\, z_n) < d(v,\, z_n).
    \end{cases}
\end{align}
Here, the hyperparameters $\alpha$ and $\beta$ represent the amplification factors. The final generated token $y_t$ is sampled from $p_\theta$. 

\subsection{Hardware and Implementtation Details}
All experiments in this paper were conducted on an NVIDIA RTX 3090 24GB GPU. For SAM, we utilized the pre-trained ViT-H SAM model. The decoding process was configured with $\alpha=1.2$ and $\beta=3$ by default. The number of objects in CVD was set to be $N*0.05$ by default.

\section{Discussion}
In this paper, we reveal that hallucinations in LVLMs primarily stem from the visual processing bottleneck caused by vision-language misalignment, which we term the visual defect. Unlike previously identified issues of language bias and statistical bias, the visual defect is challenging to resolve through high-quality data, chain-of-thought prompting, or fine-tuning, due to limitations in the pre-trained visual architecture. Based on the Information Bottleneck theory, we introduce Complementary Visual Decoupling (CVD) for visual information decoupling, incorporate a non-visual input for Non-Visual Screening (NVS) to detect hallucinations, and propose Adaptive Token-level Contrastive Decoding (ATCD) for hallucination mitigation through contrastive decoding. 

CATCH demonstrates impressive performance and robustness in perceiving fine-grained features and mitigating cumulative hallucinations in open-ended scenarios, as evidenced by the results on the CHAIR, POPE, and MME datasets across different baselines. One exciting aspect of CATCH is that it requires no additional prior knowledge, data, or training, making it efficiently applicable to various LVLMs. We hope that introducing the concept of visual defects opens new research avenues for studying hallucinations and that our complementary contrastive decoding method inspires new questions and approaches in the pretraining and fine-tuning of LVLMs.

\begin{appendices}

\end{appendices}


\bibliography{sn-bibliography}

\end{document}